\pdfoutput=1

\documentclass[11pt]{article}

\usepackage[final]{acl}

\usepackage{times}
\usepackage{latexsym}
\usepackage{graphicx}
\usepackage{url}
\usepackage{natbib}
\usepackage[svgnames]{xcolor} 

\usepackage{caption}
\usepackage{adjustbox}
\usepackage{enumitem}
\usepackage{booktabs}
\usepackage{amssymb}
\usepackage[T1]{fontenc}
\usepackage[utf8]{inputenc}

\usepackage{microtype}

\usepackage{inconsolata}

\usepackage{graphicx}

\title{A Survey of Toxicity Detection and Mitigation Strategies for Multilingual Language Models}

\author{
  \textbf{Soham Dan\textsuperscript{1}}, \textbf{Himanshu Beniwal\textsuperscript{2}},
  \textbf{Thomas Hartvigsen\textsuperscript{3}}
  \\
  \textsuperscript{1}Scale AI,
  \textsuperscript{2}Indian Institute of Technology Gandhinagar,
  \textsuperscript{3}University of Virginia
  \\
  \small{
  \texttt{soham.dan@scale.com, himanshubeniwal@iitgn.ac.in,}} 
  \small{\texttt{hartvigsen@virginia.edu}}
}

\begin{document}
\maketitle
\begin{abstract}
Large language models (LLMs) are increasingly deployed across languages, but their safety behavior remains uneven across linguistic and cultural contexts. This survey synthesizes work on toxicity detection and detoxification for multilingual LLMs. We first catalogue threat models that exploit language choice, translation pivots, code-switching, orthographic variation, multi-turn interaction, and post-deployment fine-tuning to weaken safety alignment. We then organize task formulations (toxic-to-neutral rewriting, toxicity classification, and toxic-generation evaluation), multilingual detection approaches (cross-lingual encoders, translation pipelines, representation-level probes, and LLM-based detectors), and mitigation strategies spanning data filtering, supervised and preference-based tuning, decoding-time steering, representation editing, and multilingual guardrails. Across these areas, we identify persistent challenges: uneven language coverage, culturally contingent definitions of harm, fragmented evaluation protocols, and the risk that detoxification suppresses legitimate dialectal or identity-related expression.
\end{abstract}
\begin{figure*}[t]
    \centering
    \includegraphics[width=\linewidth]{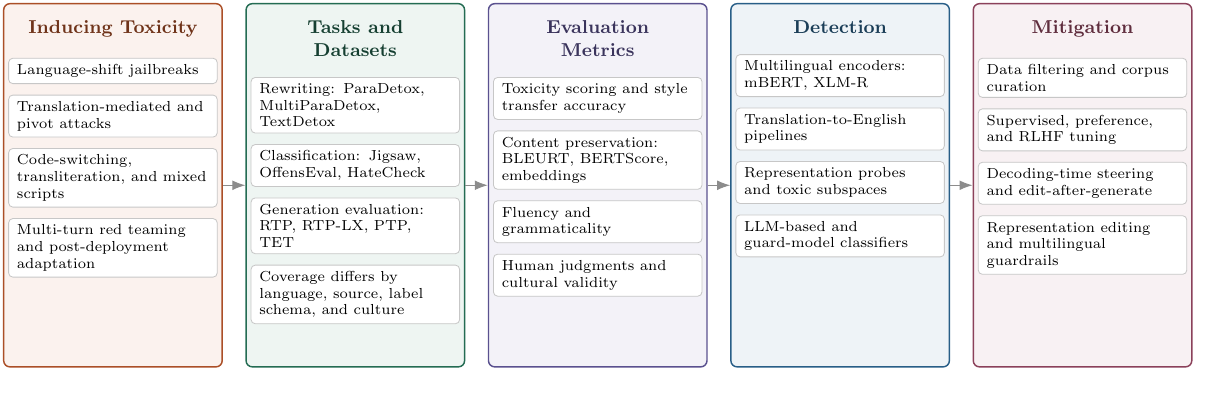}
    \caption{Taxonomy of multilingual toxicity threat models, task formulations, evaluation metrics, detection approaches, and mitigation strategies.}
    \label{fig:taxonomy}
\end{figure*}

\section{Introduction}
Large language models (LLMs) are increasingly used in multilingual settings, powering applications ranging from multilingual chatbots to cross-lingual content moderation \citep{rtplx, hartvigsen2022toxigen, kim2025decoding}. As deployment expands, so do safety risks: LLMs can produce, amplify, or fail to detect toxic content such as hate speech, harassment, profanity, and identity-based abuse, and these risks are not distributed uniformly across languages \citep{rottger2021hatecheck, sharma2025detecting, deshpande-etal-2023-chatgpt, krasnodebska-etal-2026-safety}. Despite substantial progress on English detoxification, multilingual detection and mitigation remain less mature, especially for low-resource languages, dialects, code-switched inputs, and culturally specific harms \citep{beniwal2025breakingmbadsupervisedfinetuning, ticta2021cross, multiparadetox, paradetox, rtplx}.

\paragraph{The Complexity of Multilingual Toxicity.} Multilingual detoxification is not a direct translation of English safety protocols \citep{neplenbroek2025cross, kumar2025polyguard}. Toxicity ranges from \textit{overt} categories, such as slurs, explicit insults, and profanity, to \textit{implicit} forms such as microaggressions, sarcasm, and toxic condescension, which are harder to annotate, detect, and mitigate \citep{wen2023unveiling, sap2022annotators}. Definitions of harm also vary by community: expressions that are benign, reclaimed, or dialectal in one context may be offensive in another. Multilingual settings introduce additional technical vulnerabilities. Code-switching, transliteration, and mixed-script inputs can weaken both detectors and refusal behavior \citep{zhang2023multilingual, al2024jailbreaking, yoo2025code}, while pretrained models can degenerate into toxic continuations from benign or ambiguous prompts \citep{gehman-etal-2020-realtoxicityprompts}. These failures interact with broader cultural and social biases in generated language \citep{vongpradit2024safecultural, dammu2024they}.

\paragraph{Failures of Current Approaches.} Traditional moderation systems rely heavily on keyword lists, rules, and supervised classifiers, which are brittle under paraphrase, obfuscation, dialectal variation, and context-dependent meaning \citep{kim2025decoding, huang2025content}. LLM alignment reduces many overt harms, but it does not transfer uniformly across languages: malicious prompts in lower-resource languages are more likely to elicit unsafe responses \citep{dengmultilingual, shen2024language}, and preference optimization or RLHF data remain concentrated in a small set of high-resource languages \citep{dang-etal-2024-rlhf, lu2025alignmentsafetylargelanguage}. Preference tuning can transfer across languages, but transfer quality varies with representation alignment and language-resource availability \citep{li-yong-bach-2024-preference, neplenbroek2025cross}. Machine translation is not a universal fallback either: multilingual translation systems can introduce, amplify, or obscure toxicity through hallucination and data bias \citep{costa2023toxicity}. These failures make multilingual detoxification a problem of technical robustness, evaluation validity, and sociolinguistic coverage \citep{adragna2020fairness, cecchini2024holistic}.

\par \noindent This survey provides a focused overview of detoxification for multilingual LLMs, synthesizing recent work on detection and mitigation into a taxonomy of datasets, methods, and evaluation frameworks (Figure~\ref{fig:taxonomy}). Related surveys examine multilingual LLM safety broadly \citep{krasnodebska-etal-2026-safety}; our emphasis is the narrower detoxification pipeline: how toxic behavior is induced, measured, detected, and mitigated across languages.
\paragraph{Scope and Contributions.} The survey is organized around the following themes:
\begin{itemize}[nosep]
    \item We organize multilingual threat models covering language-shift jailbreaks, translation/pivot attacks, code-switch prompts, multilingual red-teaming, and adaptation-time safety collapse from cross-lingual fine-tuning (\S\ref{sec:threat-models}).
    \item We organize task formulations into three categories---toxic-to-neutral rewriting, toxicity classification, and toxic-generation/prompt continuation---and survey the datasets and metrics used to evaluate each.
    \item We survey multilingual toxicity detection methods, spanning encoder- and decoder-based transformers, translation-based pipelines, representation-level probing, and LLM-based zero-shot detection.
    \item We present a mechanism-based detoxification taxonomy covering data-centric filtering, supervised and preference-based tuning, decoding-time steering, representation editing, and multilingual guardrails.
\end{itemize}

\noindent We conclude with a discussion of open challenges---cross-lingual coverage gaps, cultural misalignment, evaluation fragmentation, and over-suppression---and identify concrete directions for building globally safe and equitable LLMs.

\section{Threat Models for Inducing Toxicity in Multilingual LLMs}
\label{sec:threat-models}

We focus on \emph{multilingual-specific} toxicity-inducing threat models, which are adversarial procedures that exploit language choice, cross-lingual transfer, or multilingual interaction to elicit toxic outputs from a safety-aligned model.
In this survey, we treat safety vulnerabilities---jailbreaks, alignment bypass, red-teaming---as the mechanisms through which toxic outputs are induced; ``safety failure'' and ``toxicity elicitation'' are thus two views of the same problem.
To make these threat models comparable, we use four diagnostic axes: \textbf{(i)} language composition (monolingual vs.\ code-switched), \textbf{(ii)} script composition (standard vs.\ mixed-script/transliterated), \textbf{(iii)} translation mediation (direct vs.\ pivot/round-trip), and \textbf{(iv)} cultural-norm variation (universal vs.\ culturally contingent harm).
The subsections below instantiate these axes: language-shift attacks isolate non-English prompting; translation-mediated attacks stress pivoting and round-trip evaluation; code-switching attacks combine language and script composition; and multilingual red-teaming/adaptation attacks expose how these linguistic operators interact with culturally contingent safety policies.

\subsection{Prompt-Space Multilingual Attacks}
\label{sec:prompt-attacks}

\paragraph{Language-Shift Jailbreaks.}
This threat primarily tests \emph{language composition}: a malicious or ambiguous request remains monolingual but is re-expressed outside English. \citet{dengmultilingual} formalize (i) \emph{unintentional} multilingual jailbreaks (benign users prompting in underrepresented languages) and (ii) \emph{intentional} multilingual jailbreaks (adversaries combining multilingual prompts with explicit malicious instructions), and show substantially higher unsafe rates in lower-resource languages.

\paragraph{Translation-Mediated and Pivot Attacks.}
This threat stresses \emph{translation mediation}: an unsafe English prompt is translated into a target low-resource language to increase compliance, then the response is translated back. \citet{shen2024language} empirically demonstrate higher unsafe response rates for malicious prompts expressed in lower-resource languages, motivating translation/pivot-based red-teaming. Recent defenses that \emph{re-anchor} safety using English while enforcing target-language outputs further underscore translation as a core failure mode in multilingual safety \citep{zhang2025english}.

\paragraph{Language Mixing: Code-Switching and Multi-Language Mixtures.}
This threat combines \emph{language composition} with \emph{script composition}, because multilingual prompts may mix languages, scripts, and transliterated forms within one context. \citet{yoo2025code} show that code-switched red-teaming queries can elicit unsafe behavior more effectively than monolingual attacks and introduce a synthesis framework (CSRT) to generate such queries at scale. Complementarily, \citet{upadhayay2024sandwich} propose the \emph{Sandwich Attack}, a multi-language mixture prompt that interleaves benign and adversarial segments across languages to induce harmful completions in a black-box setting.

\subsection{Multilingual Red Teaming}
\label{sec:multilingual-redteaming}

Red teaming operationalizes these axes by generating adversarial prompts and dialogues at scale, including culturally specific prompts whose harmfulness may not be captured by English-centric policies. Early work established manual and LM-assisted red teaming methodologies \citep{perez2022red, zhuo2023red}. Recent multilingual extensions explicitly target the multilingual capability envelope: CSRT generates code-switched attacks \citep{yoo2025code}; Rainbow Teaming produces diverse open-ended adversarial prompts and has been replicated/extended for Polish as a concrete non-English safety stress test \citep{samvelyan2024rainbow, krasnodkebska2025rainbow}; and MM-ART automates \emph{multi-turn, multilingual} red teaming, showing vulnerability increases sharply with conversation length and is substantially underestimated by single-turn English evaluation \citep{singhania2025multi}.

\subsection{Post-Deployment Adaptation Attacks}
\label{sec:adaptation-attacks}

\paragraph{Cross-lingual Fine-Tuning Attacks.}
Aligned multilingual models are frequently customized via SFT/PEFT after deployment, creating an adaptation-time attack surface where safety behavior can shift across languages and local norms. \citet{poppi2025towards} show that fine-tuning on a small toxic dataset in \emph{one} language can collapse safety across \emph{other} languages (cross-lingual attack transfer). Their Safety Information Localization (SIL) analysis suggests safety-relevant parameters are partially language-agnostic, enabling sparse updates to induce multilingual failure.

\paragraph{Jailbreaks via New-Language Learning.}
Even benign adaptation can be risky: \citet{upadhayay-behzadan-2025-tongue} show that LoRA fine-tuning to learn a low-resource language---without harmful data---can nonetheless degrade refusal behavior, implying that multilingual expansion itself can destabilize safety guarantees.

Multilingual detoxification methods should therefore be evaluated not only on monolingual English prompts, but under compositions of multilingual operators (translate/pivot, code-switch, mixture prompts, transliteration), multi-turn interaction, and post-deployment adaptation stress tests. The threat models above motivate the task formulations, datasets, and metrics we discuss next.

\section{Task Setup: Datasets and Metrics}
\subsection{Datasets}
Toxicity datasets can broadly be categorized into three tasks, each corresponding to a distinct evaluation goal:
\par \noindent \textbf{Toxic-to-Neutral Rewriting.}
ParaDetox \citep{paradetox} introduced more than 10K English toxic$\rightarrow$neutral paraphrase pairs. Subsequent work explored cross-lingual transfer for detoxification \citep{moskovskiy-etal-2022-exploring, dementieva-etal-2023-exploring}, added a Hindi evaluation set \citep{sourabrata-etal-2023-text}, and extended the ParaDetox collection pipeline to Russian, Ukrainian, and Spanish in MultiParaDetox \citep{multiparadetox}. The TextDetox/PAN 2024 shared task and its COLING extension broadened the parallel detoxification setting to 9 languages: English, Spanish, German, Chinese, Arabic, Hindi, Ukrainian, Russian, and Amharic \citep{dementieva2024overview, dementieva-etal-2025-multilingual}. SynthDetox-M \citep{moskovskiy2025synthdetoxm, moskovskiy-etal-2024-llms} adds 16K high-quality synthetic pairs across German, French, Spanish, and Russian via few-shot LLM prompting. APPDIA \citep{atwell2022appdia} and CAPP \citep{som-etal-2024-demonstrations} provide discourse- or dialogue-aware parallel corpora for offensive-content paraphrasing.

 \par \noindent \textbf{Toxic Text Detection.}
Jigsaw's English Toxic Comment and Unintended Bias tasks provide large-scale comment-level toxicity labels, while the multilingual Jigsaw task evaluates binary toxicity in Spanish, Italian, Turkish, French, Portuguese, and Russian using English-labeled training data \citep{jigsaw-toxic-comment-classification-challenge, jigsaw-multilingual-toxic-comment-classification}. OffensEval covers English offensive-language identification in 2019 and five languages in 2020 (Arabic, Danish, English, Greek, and Turkish) \citep{zampieri2019semeval, zampieri-etal-2020-semeval}; the related Toxic Spans task targets span-level explanations in English \citep{pavlopoulos-etal-2021-semeval}. HateCheck provides functional tests for English hate-speech detection, and Multilingual HateCheck extends this diagnostic framing to ten languages \citep{rottger2021hatecheck, rottger2022multilingual}. HASOC and HatEval provide additional multilingual hate/offensive-language benchmarks \citep{mandl2019overview, basile2019semeval}. LifeTox \citep{kim2024lifetox} targets implicit toxicity in English advice-seeking contexts, and ToxiGen \citep{hartvigsen2022toxigen} provides 274K machine-generated toxic and benign statements about protected groups. Such classification datasets serve both toxicity evaluation \citep{koh-etal-2024-llms} and retrieval-based detoxification \citep{pozzobon-etal-2023-goodtriever}.

\par \noindent \textbf{Non-Toxic Text Continuation.}
RealToxicityPrompts (RTP) \citep{gehman-etal-2020-realtoxicityprompts} provides 100K English web prompts scored by Perspective API and introduced common toxic-generation metrics such as Expected Maximum Toxicity (EMT) and toxicity probability. RTP-LX \citep{rtplx} extends this style of evaluation to 28 languages in the paper, with human-transcreated prompts and native-speaker annotations for harm categories such as bias, insult, identity attack, and microaggression. PolygloToxicityPrompts (PTP) \citep{jain-etal-2024-polygloToxicityprompts} offers 425K naturally sourced prompts across 17 languages and reports that toxicity tends to increase as model size grows or language-resource availability decreases. FrenchToxicityPrompts \citep{brun2024frenchtoxicityprompts} provides 50K French prompts. TET \citep{luong2024realistic} comprises 2,546 prompts filtered from 1M real-world LLM interactions to expose latent toxic behaviors that can bypass safety mechanisms. \citet{deshpande-etal-2023-chatgpt} further showed that persona-based system prompts can amplify toxic degeneration.

\begin{table*}[t]
\centering
\small
\begin{adjustbox}{max width=\textwidth}
\begin{tabular}{llclr}
\toprule
\textbf{Dataset} & \textbf{Task} & \textbf{Languages} & \textbf{Source} & \textbf{Approx. size} \\
\midrule
ParaDetox \citep{paradetox} & Rewrite & EN & Natural & 10K+ \\
MultiParaDetox \citep{multiparadetox} & Rewrite & RU, UK, ES & Natural & 16.4K pairs \\
SynthDetox-M \citep{moskovskiy2025synthdetoxm} & Rewrite & DE, FR, ES, RU & Synthetic & 16K \\
TextDetox/PAN 2024 \citep{dementieva2024overview} & Rewrite & 9 langs & Mixed & 1K pairs/lang \\
APPDIA \citep{atwell2022appdia} & Rewrite & EN (Reddit) & Natural & $\sim$2K \\
\midrule
Jigsaw \citep{jigsaw-toxic-comment-classification-challenge,jigsaw-multilingual-toxic-comment-classification} & Classify & EN + 6 eval langs & Natural & $\sim$160K EN train \\
OffensEval \citep{zampieri2019semeval,zampieri-etal-2020-semeval} & Classify & 5 langs & Natural & Task-dependent \\
HateCheck \citep{rottger2021hatecheck,rottger2022multilingual} & Classify & EN + 10 langs & Synthetic tests & 40K+ tests \\
LifeTox \citep{kim2024lifetox} & Classify & EN & Natural & 87.5K \\
ToxiGen \citep{hartvigsen2022toxigen} & Classify & EN & Synthetic & 274K \\
\midrule
RTP \citep{gehman-etal-2020-realtoxicityprompts} & Generate & EN & Natural & 100K \\
RTP-LX \citep{rtplx} & Generate & 28 langs (paper) & Transcreated & 1K+/locale \\
PTP \citep{jain-etal-2024-polygloToxicityprompts} & Generate & 17 langs & Natural & 425K \\
FrenchTP \citep{brun2024frenchtoxicityprompts} & Generate & FR & Natural & 50K \\
TET \citep{luong2024realistic} & Generate & EN & Natural & 2,546 \\
\bottomrule
\end{tabular}
\end{adjustbox}
\caption{Taxonomy of toxicity datasets organized by task. \textit{Source} indicates whether data is human-authored (Natural), machine-translated (Translated), human-transcreated (Transcreated), LLM-generated (Synthetic), or assembled from multiple sources (Mixed). Sizes are rounded where appropriate and follow the cited paper or task release.}
\label{tab:dataset_taxonomy}
\end{table*}

\subsection{Metrics}
 \par \noindent \textbf{Toxicity Detection Metrics}
Outputs are often scored by toxicity classifiers. A common metric is \textbf{\textit{style transfer accuracy (STA)}}: the fraction of outputs that a classifier deems non-toxic. For example, models use RoBERTa-based classifiers trained on Jigsaw to compute STA \citep{dementieva-etal-2023-exploring}. Other tools such as the \textbf{\textit{Perspective API}}\footnote{\url{https://perspectiveapi.com/}} provide continuous toxicity scores. Detoxification systems are typically expected to improve STA or reduce toxicity scores while preserving meaning and fluency (e.g., reducing toxic-generation probability as in \citealp{li-yong-bach-2024-preference}).

 \par \noindent \textbf{Content Preservation and Fluency} To ensure meaning is retained, similarity metrics are applied. Popular choices include \textbf{\textit{BLEURT}} \citep{sellam2020bleurt} or \textbf{\textit{BERTScore}} \citep{zhang-etal-2020-bertscore} to compare the detoxified output to the input or a reference. \citet{dementieva-etal-2023-exploring} adopt BLEURT for English content similarity (SIM) and LaBSE embeddings for Russian. \textbf{\textit{Fluency}} is evaluated by the percentage of grammatical or fluent sentences, often via a language acceptability classifier (e.g., a RoBERTa trained to recognize acceptability) \citep{dementieva-etal-2023-exploring, paradetox}. Combined metrics like the product of STA, SIM, and fluency are sometimes used to rank models.

\noindent \par \textbf{Cross-Lingual Alignment}
When detox and translation happen together, one can also measure translation quality or cross-lingual consistency. For example, in simultaneous translation+detox, one may compute BLEU \citep{papineni-etal-2002-bleu} or COMET \citep{rei-etal-2020-comet} between the generated detoxified translation and a human reference.
In practice, cross-lingual transfer effectiveness is often inferred from zero-shot performance, or by correlating translated and original outputs. Some work also uses source-output embedding similarity as a proxy for semantic alignment.

\noindent \par \textbf{Human Evaluation}
Ultimately, manual judgments are key. Human annotators typically rate detox outputs on \textbf{\textit{(1)}} toxicity/style (is the output non-toxic/neutral?); \textbf{\textit{(2)}} content preservation (does it retain the original meaning?); and \textbf{\textit{(3)}} fluency (is the output natural?). Human scores are used both to evaluate final systems and to calibrate or validate automatic metrics (e.g. correlating BLEURT with meaning preservation).

\section{Detection}
Detecting toxicity in multilingual settings is complicated by linguistic diversity, code-mixing, dialectal variation, and culturally contingent definitions of harm.

\subsection{Multilingual Transformers}
Early toxicity detection relied on keyword lists and lexicon-based classifiers, which lack contextual understanding and fail under paraphrase, obfuscation, and dialectal variation. Deep contextual encoders such as mBERT and XLM-R marked a significant advance, demonstrating that cross-lingual representations can improve toxicity identification across languages \citep{conneau2020unsupervised, ticta2021cross}. These models benefit from shared subword vocabularies and multilingual pretraining, allowing transfer from high-resource languages such as English to languages with less labeled toxicity data. Nevertheless, performance remains uneven across scripts, dialects, and languages with limited pretraining resources \citep{kanjirangat2025tokenization}. The brittleness of subword tokenization under spelling variants, obfuscation, and script mixing has motivated byte- and character-level alternatives; the newer Perspective API, for example, uses a multilingual token-free Charformer architecture for toxic-content detection \citep{lees2022new}.

\subsection{Translation-Based Pipelines}
A parallel line of work explores translation-based pipelines, where non-English text is machine-translated into English before being passed to an English toxicity classifier \citep{bell2025translate}. This strategy can be competitive because English detectors are comparatively mature, but it introduces error propagation, translation artifacts, and semantic drift, especially for dialectal, code-mixed, or morphologically complex inputs \citep{zampieri-etal-2020-semeval}. Translation systems themselves can introduce or obscure toxic content, so translation is best treated as an evaluation or deployment design choice rather than a neutral preprocessing step \citep{costa2023toxicity}.

\subsection{Representation-Level Detection}
Recent research identifies linear toxic subspaces in language model embeddings \citep{wang2021simple, duan2025gloss}, suggesting that toxicity-related features can occupy identifiable directions in latent space. Decomposing models into interpretable expert components can further isolate toxicity-related behavior \citep{shaik2025redefining}. These findings motivate probing and attribution techniques that seek to locate where toxicity features are stored, with applications to both detection and mitigation \citep{safedit, goyal2025breaking}.

\subsection{LLM-Based Detection}
The emergence of instruction-tuned LLMs has opened new detection avenues \citep{hu2024toxicity}. Several works evaluate LLMs as zero-shot or few-shot toxicity detectors, showing strong generalization but also calibration failures \citep{liucalibration} and cultural misalignment across languages \citep{yang-etal-2025-mrguard}. These models often rely on implicit safety priors learned during alignment, which can produce inconsistent behavior on region-specific sociolinguistic norms.

\par \noindent \textbf{\textit{Takeaway}}: Multilingual LLMs and multilingual encoders have improved cross-lingual toxicity detection, but substantial challenges remain. Persistent gaps in language coverage, bias in training corpora, inconsistent cross-lingual performance, and translation-induced errors limit the reliability of current detectors. See Table~\ref{tab:detection_comparison} in the Appendix for a detailed comparison.

\section{Detoxification}
\subsection{Data-Centric Detoxification}
Data-centric detoxification targets the quality of pre-training and fine-tuning corpora by removing or down-weighting toxic content. Early filtering pipelines relied on blocklists or lexical heuristics; contemporary pipelines often combine language identification, quality filters, and toxicity classifiers at web scale \citep{Kreutzer_2022, stranisci2025they}. More recent work emphasizes bias-aware filtering to avoid suppressing dialectal or marginalized speech \citep{sap2022annotators, jaggi-etal-2024-accurate, xu2021detoxifying}. In multilingual settings, filtering depends heavily on cross-lingual detector generalization, which can misclassify culturally specific idioms, reclaimed slurs, or dialectal markers \citep{bensalem2024toxiclanguagedetectionsystematic, welbl2021challenges}. Data filtering is scalable and can reduce exposure to toxic training examples, but it also risks cultural misalignment, uneven language coverage, and the over-removal of minority language varieties; recent work argues that harmful-content filtering can deepen underrepresentation of already vulnerable groups \citep{stranisci2025they}.

\subsection{Model-Centric Detoxification}
\paragraph{Supervised Finetuning on Safe or Contrastive Data} Supervised detoxification approaches fine-tune LLMs on curated non-toxic corpora, contrastive toxic--neutral pairs, or attribute-controlled toxicity objectives \citep{hawkins2024effect, Meng2024}. \citet{neplenbroek2025cross} report that mitigation can transfer across languages, but that transfer depends on language-resource conditions and can trade off against non-English generation quality.
Fine-tuning-based detoxification can provide strong control, but it may reduce output diversity, degrade generation quality, or introduce stylistic flattening \citep{wang2022exploring, welbl2021challenges}.

\paragraph{Instruction-Based Safety Tuning}
Instruction tuning using curated safety data or synthetic refusal-style instructions can enhance multilingual LLMs' ability to decline harmful requests and avoid toxic continuations. Multilingual preference optimization shows that alignment can transfer across languages when feedback data are balanced and sufficiently broad \citep{dang-etal-2024-rlhf}. These methods scale well for deployment, though annotation biases and cultural coverage remain persistent limitations.

\paragraph{RLHF and Human Feedback Alignment} Reinforcement learning from human feedback (RLHF) \citep{ouyang2022training, bai2022constitutionalai} can improve safety by training reward models to penalize toxic outputs. While RLHF datasets are primarily English-centric, multilingual LLMs can benefit indirectly through shared parameters and cross-lingual transfer \citep{dang-etal-2024-rlhf}. However, reliance on English safety norms introduces cross-cultural misalignment in multilingual models \citep{lu2025alignmentsafetylargelanguage}, especially for expressions that are offensive in some cultures but neutral in others.

\subsection{Decoding-Time Detoxification}
Post-hoc methods avoid or minimize retraining by steering generation at inference \citep{ko2024large}. Classifier-guided and expert-based logit steering include PPLM hidden-state perturbations \citep{dathathri2020plug, pascual-etal-2021-plug-play}, GeDi-style generative discriminators \citep{krause2021gedi}, and expert/anti-expert mixture decoding such as DExperts \citep{liu-etal-2021-dexperts}. Expert steering is modular, but high-quality multilingual experts are a bottleneck. A second family uses \emph{edit-after-generate}: produce a candidate, detect toxicity, and rewrite or refine it via prompting or a specialized editor \citep{leong2023self}. In multilingual deployments, \emph{translation-pivot pipelines} (translate$\rightarrow$detox in English$\rightarrow$translate back) remain common, but they risk semantic drift and can erase culturally salient pragmatics \citep{dementieva-etal-2023-exploring, bell2025translate}. Retrieval augmentation can also support detoxification by grounding rewrites in policy examples or safe templates \citep{pozzobon-etal-2023-goodtriever}.

\subsection{Model Editing and Representation Interventions}
Recent work investigates activation steering: modifying internal LM representations to remove or attenuate toxic features \citep{goyal2025breaking}. Activation Addition \citep{turner2024steeringlanguagemodelsactivation} and ROME-based editing \citep{meng2022locating} identify directions or associations that can be altered during generation. Early analyses of how interventions reshape cross-lingual representations \citep{sundar2025steering} suggest potential for multilingual transfer, though evaluation is still nascent and regression risk remains high without careful cross-lingual audits \citep{safedit}.

\subsection{Multilingual Guardrails}
A related line of work--not the main focus of this survey--is post-hoc moderation via multilingual guardrails \citep{yi2024position}: deployment-time controllers that classify and gate prompts and responses into policy categories such as prompt harmfulness, response harmfulness, and refusal/compliance under adversarial multilingual inputs. Language choice, code-switching, and transliteration can weaken English-centric safeguards. Representative guardrails and safety classifiers include Llama Guard \citep{inan2023llama}, Aegis \citep{ghosh2024aegis}, MrGuard \citep{yang-etal-2025-mrguard}, WildGuard \citep{han2024wildguard}, PolyGuard \citep{kumar2025polyguard}, MultiGuard/OmniGuard \citep{verma2025multiguard}, CREST \citep{bansal2025crest}, Qwen3Guard \citep{zhao2025qwen3guard}, and UnityAI-Guard \citep{beniwal2025unityaiguardpioneeringtoxicitydetection}.

\par \noindent \textbf{\textit{Key Takeaways.}}
\begin{itemize}
    \item Cross-lingual robustness remains a central challenge: Detoxification methods often perform better in high-resource languages than in low-resource or morphologically rich languages.
    \item Cultural bias persists across detoxification pipelines: Much safety supervision originates from English, creating misalignment in non-Western contexts.
    \item Hybrid strategies are promising: Combining data filtering, controlled decoding, alignment tuning, and guardrails can cover failure modes that no single method handles reliably.
    \item Avoiding over-censorship is an unresolved issue: Techniques often suppress legitimate emotional or dialectal expressions, leading to ``model homogenization.''
\end{itemize}
\noindent See Table~\ref{tab:detox_techniques} in the Appendix for a detailed comparison of detoxification techniques.

\section{Discussion and Open Challenges}
\subsection{Cross-Lingual Gaps in Detoxification}
A persistent disparity exists between high-resource and low-resource languages. Many multilingual toxicity detectors and safety-tuned LLMs are trained or validated primarily on English and other high-resource languages, leaving morphologically rich, dialectal, or culturally distant varieties under-detected \citep{shen2024language, wang2024all, bensalem2024toxiclanguagedetectionsystematic}. Alignment methods such as RLHF and constitutional tuning have historically relied on English-heavy preference or principle data \citep{ouyang2022training, bai2022constitutionalai}, which can produce inconsistent refusal behavior and weak recognition of non-English toxic slang \citep{lu2025alignmentsafetylargelanguage, dang-etal-2024-rlhf}.

\par \noindent \textbf{\textit{Open Challenge}}: Developing culturally aware multilingual safety representations that scale to low-resource languages without English over-dominance remains essential.

\subsection{Cultural and Normative Misalignment}
Toxicity is culturally embedded: annotators' identities and beliefs strongly influence judgments \citep{sap2022annotators, sap-etal-2019-risk, jaggi-etal-2024-accurate}, yet many safety datasets collapse disagreement into a single label. Models therefore risk over-censoring reclaimed slurs, misclassifying dialectal expressions, or reinforcing majority-group norms \citep{shen2024language}. Languages with rich honorific systems, code-switching norms, or culturally specific humor \citep{li2024culturellm} expose current models' limited ability to differentiate toxicity from socially sanctioned expression.

\par \noindent \textbf{\textit{Open Challenge}}: Future systems need culturally grounded, community-driven annotation and context-aware toxicity modeling that respects sociolinguistic diversity.

\subsection{Lack of Robust, Multilingual Evaluation Frameworks}
A recurring theme is the lack of standardized, multilingual frameworks for evaluating toxicity. Existing generation benchmarks such as RealToxicityPrompts \citep{gehman-etal-2020-realtoxicityprompts} are English-only, while newer multilingual datasets such as RTP-LX, PTP, and PolyGuard broaden coverage but differ substantially in task format, label schema, and language set \citep{rtplx, jain-etal-2024-polygloToxicityprompts, kumar2025polyguard}. Evaluation pipelines also struggle with subtle harms such as microaggressions, presuppositional harm, and implicit bias \citep{sap2022annotators}. Cross-lingual transfer of toxicity classifiers can produce false positives for dialects or false negatives for low-resource slang, making direct comparison unreliable.

\par \noindent \textbf{\textit{Open Challenge}}: The field needs multilingual benchmarks with fine-grained toxicity categories, cross-cultural annotations, and shared evaluation protocols \citep{wang2024all}.

\subsection{Over-Suppression and Style Degradation}
Detoxification techniques, particularly contrastive finetuning and representation editing, can reduce linguistic richness or stylistic diversity. Prior work shows that detoxification can trade toxicity reduction for reduced fluency, reduced diversity, or suppression of identity-related language \citep{welbl2021challenges, liu-etal-2021-dexperts, xu2021detoxifying}. In multilingual settings, this risk is amplified: low-resource languages may be pushed toward generic, formal, or English-like outputs because the model has weaker language-specific representations. Techniques such as activation editing \citep{turner2024steeringlanguagemodelsactivation} and PPLM \citep{dathathri2020plug} offer fine-grained control but still risk semantic over-suppression when applied cross-lingually.

\par \noindent \textbf{\textit{Open Challenge}}: Designing detoxification techniques that preserve stylistic and cultural characteristics while eliminating harmful content remains an open frontier.

\subsection{Handling Code-Switching and Mixed-Linguistic Toxicity}
Multilingual communities frequently communicate through code-switching (e.g., Hinglish, Arabizi, Spanglish), combining scripts, phonetic spellings, and culturally specific expressions. Current LLMs and toxicity detectors are less reliable under code-switching because training coverage, tokenization, and evaluation data are sparse for mixed-language inputs \citep{zhang2023multilingual, bensalem2024toxiclanguagedetectionsystematic}. Safety failures under code-switched or transliterated prompts have been demonstrated for red-teaming and jailbreak settings \citep{al2024jailbreaking, yoo2025code}, and Hindi-English toxic language remains an active detection problem \citep{sharma2025detecting}. This poses serious risks for global deployments of multilingual LLMs.

\par \noindent \textbf{\textit{Open Challenge}}: Robust multilingual safety systems must explicitly account for code-switching and orthographic variation via code-mixed training corpora, unified mixed-script tokenizers, and transliteration-aware detection.

\subsection{The Role of Decoding-Time Steering}
Decoding-time methods (PPLM, GeDi, DExperts) are best viewed as a \emph{complementary} safeguard---modular and retraining-light---not a standalone fix for root causes like English-centric alignment data. In multilingual settings, tokenization asymmetries, script mixing, and cross-lingual semantic drift weaken expert model reliability; building language-specific experts for low-resource languages remains impractical at scale.
A central bottleneck is \textbf{expert availability} (data scarcity), followed by \textbf{representation entanglement} (toxic directions conflating sentiment, intensity, and identity) and cross-script transfer instability.
The evidence points toward \textbf{hybrid architectures}: filtering and alignment tuning address root causes; steering provides inference-time control; guardrails add system-level robustness. Where norms are culturally contingent, community-grounded supervision remains necessary.

\subsection{Key Takeaways}
\begin{itemize}[nosep]
    \item \textbf{Language disparities}: Methods effective in English often underperform in low-resource languages and dialects.
    \item \textbf{Cultural context}: One-size-fits-all safety tuning misaligns with local norms, over-censoring benign expressions or missing contextually offensive language.
    \item \textbf{Evaluation gaps}: Fragmented protocols and English-centric benchmarks make cross-system comparison unreliable, especially for subtle toxicity.
    \item \textbf{Style trade-offs}: Detoxification often degrades output diversity, yielding generic text that erases linguistic richness.
    \item \textbf{Hybrid approaches}: Combining data filtering, controlled generation, culturally aware alignment, and guardrails is the most defensible direction for deployment.
    \item \textbf{Interpretability}: Understanding why models flag or generate toxic content is essential for trust and auditability in multilingual settings.
\end{itemize}

\section{Conclusion}
This survey offers a focused treatment of detoxification for multilingual LLMs, a problem that remains under-studied relative to its practical importance.
We systematized the space along three axes: multilingual threat models that expose how language shift, translation pivots, code-switching, and post-deployment adaptation erode safety; task formulations spanning rewriting, classification, and toxic-generation evaluation; and a mechanism-based taxonomy covering data filtering, supervised and preference-based tuning, decoding-time steering, representation editing, and guardrails.

Two findings cut across every axis. First, cross-lingual transfer of safety is unreliable: methods effective in English routinely under-perform in low-resource and morphologically rich languages, and alignment learned from English preference data can misfire when projected onto other cultural contexts. Second, detoxification and linguistic diversity are in tension: current techniques can suppress legitimate dialectal, code-switched, or identity-related expression, trading one harm for another.

The most pressing research need is evaluation infrastructure: standardized, culturally grounded multilingual benchmarks that go beyond English-translated prompts and that measure not only toxicity reduction but also preservation of stylistic and cultural content. Without such benchmarks, progress on multilingual safety will remain difficult to measure and easy to overstate.

\section*{Limitations}

This survey synthesizes a fast-moving literature, so specific model families, benchmarks, and best practices may evolve after publication. Its scope is also intentionally focused on text-based toxicity detection and detoxification for multilingual language models; we do not cover multimodal moderation, broader cyber-safety policies, or legal governance in depth. The evidence base is uneven across languages: many ``multilingual'' studies still emphasize English and other high-resource languages, with fewer results for low-resource languages, dialect continua, and code-mixed or transliterated text. Because toxicity definitions and label schemas vary across datasets and cultures, comparisons across papers are necessarily approximate. We also do not run a quantitative meta-analysis or reproduce prior experiments; our synthesis depends on reported results, which often use different models, datasets, detectors, and evaluation protocols. Finally, many evaluations rely on automatic detectors, translation-based protocols, or closed-model assessments, which can introduce measurement noise and limit strict apples-to-apples replication.

\section*{Ethics}
This survey reviews prior work on toxicity in multilingual language models and does not involve new data collection, human-subject annotation, or model deployment. Because the paper discusses jailbreaks, red-teaming, and adaptation-time safety failures, the topic has some dual-use risk. We therefore keep the discussion at the level of threat models, evaluation categories, and mitigation strategies rather than providing operational attack instructions or harmful prompt examples. The central ethical concern is that automated toxicity detection and detoxification can reflect English-centric or majority-culture norms, misclassify reclaimed or dialectal expressions, and suppress legitimate identity-related speech. Such failures can reinforce societal and annotator biases or lead to over-censorship, especially for communities already underrepresented in training and evaluation data. We therefore emphasize culturally grounded evaluation, inclusive data practices, transparent reporting of language coverage, and careful safety--utility trade-offs in multilingual deployment.
\bibliography{custom}

@inproceedings{dementieva-etal-2025-multilingual,
    author    = {Daryna Dementieva and Nikolay Babakov and Amit Ronen and Abinew Ali Ayele and Naquee Rizwan and Florian Schneider and Xintong Wang and Seid Muhie Yimam and Daniil Alekhseevich Moskovskiy and Elisei Stakovskii and Eran Kaufman and Ashraf Elnagar and Animesh Mukherjee and Alexander Panchenko},
    title     = {Multilingual and Explainable Text Detoxification with Parallel Corpora},
    booktitle = {Proceedings of the 31st International Conference on Computational Linguistics},
    pages     = {7998--8025},
    address   = {Abu Dhabi, UAE},
    publisher = {Association for Computational Linguistics},
    year      = {2025},
    url       = {https://aclanthology.org/2025.coling-main.535/},
    note      = {COLING 2025} 
}

@inproceedings{dengmultilingual,
  title={Multilingual Jailbreak Challenges in Large Language Models},
  author={Deng, Yue and Zhang, Wenxuan and Pan, Sinno Jialin and Bing, Lidong},
  booktitle={The Twelfth International Conference on Learning Representations},
  year={2024},
  url={https://proceedings.iclr.cc/paper_files/paper/2024/hash/6b396f766a50e0853a5164e68048540c-Abstract-Conference.html}
}

@inproceedings{upadhayay2024sandwich,
  title={Sandwich attack: Multi-language Mixture Adaptive Attack on LLMs},
  author={Upadhayay, Bibek and Behzadan, Vahid},
  booktitle={Proceedings of the 4th Workshop on Trustworthy Natural Language Processing (TrustNLP 2024)},
  month={jun},
  address={Mexico City, Mexico},
  publisher={Association for Computational Linguistics},
  pages={208--226},
  year={2024},
  url={https://aclanthology.org/2024.trustnlp-1.18/},
  doi={10.18653/v1/2024.trustnlp-1.18}
}

@inproceedings{singhania2025multi,
  title={Multi-lingual multi-turn automated red teaming for LLMs},
  author={Singhania, Abhishek and Dupuy, Christophe and Mangale, Shivam Sadashiv and Namboori, Amani},
  booktitle={Proceedings of the 5th Workshop on Trustworthy NLP (TrustNLP 2025)},
  month={may},
  address={Albuquerque, New Mexico},
  publisher={Association for Computational Linguistics},
  pages={141--154},
  year={2025},
  url={https://aclanthology.org/2025.trustnlp-main.11/},
  doi={10.18653/v1/2025.trustnlp-main.11}
}

@inproceedings{perez2022red,
  title={Red Teaming Language Models with Language Models},
  author={Perez, Ethan and Huang, Saffron and Song, Francis and Cai, Trevor and Ring, Roman and Aslanides, John and Glaese, Amelia and McAleese, Nat and Irving, Geoffrey},
  booktitle={Proceedings of the 2022 Conference on Empirical Methods in Natural Language Processing},
  month={dec},
  address={Abu Dhabi, United Arab Emirates},
  publisher={Association for Computational Linguistics},
  pages={3419--3448},
  year={2022},
  url={https://aclanthology.org/2022.emnlp-main.225/},
  doi={10.18653/v1/2022.emnlp-main.225}
}

@article{samvelyan2024rainbow,
  title={Rainbow teaming: Open-ended generation of diverse adversarial prompts},
  author={Samvelyan, Mikayel and Raparthy, Sharath C and Lupu, Andrei and Hambro, Eric and Markosyan, Aram H and Bhatt, Manish and Mao, Yuning and Jiang, Minqi and Parker-Holder, Jack and Foerster, Jakob},
  journal={Advances in Neural Information Processing Systems},
  volume={37},
  pages={69747--69786},
  year={2024},
  url={https://proceedings.neurips.cc/paper_files/paper/2024/hash/8147a43d030b43a01020774ae1d3e3bb-Abstract-Conference.html},
  doi={10.52202/079017-2229}
}

@inproceedings{krasnodkebska2025rainbow,
  title={Rainbow-Teaming for the Polish Language: A Reproducibility Study},
  author={Krasnod{\k{e}}bska, Aleksandra and Chrabaszcz, Maciej and Kusa, Wojciech},
  booktitle={Proceedings of the 5th Workshop on Trustworthy NLP (TrustNLP 2025)},
  month={may},
  address={Albuquerque, New Mexico},
  publisher={Association for Computational Linguistics},
  pages={155--165},
  year={2025},
  url={https://aclanthology.org/2025.trustnlp-main.12/},
  doi={10.18653/v1/2025.trustnlp-main.12}
}

@inproceedings{shen2024language,
  title={The Language Barrier: Dissecting Safety Challenges of LLMs in Multilingual Contexts},
  author={Shen, Lingfeng and Tan, Weiting and Chen, Sihao and Chen, Yunmo and Zhang, Jingyu and Xu, Haoran and Zheng, Boyuan and Koehn, Philipp and Khashabi, Daniel},
  booktitle={Findings of the Association for Computational Linguistics: ACL 2024},
  month={aug},
  address={Bangkok, Thailand},
  publisher={Association for Computational Linguistics},
  pages={2668--2680},
  year={2024},
  url={https://aclanthology.org/2024.findings-acl.156/},
  doi={10.18653/v1/2024.findings-acl.156}
}

@inproceedings{luong2024realistic,
  title={Realistic evaluation of toxicity in large language models},
  author={Luong, Tinh and Le, Thanh-Thien and Ngo, Linh and Nguyen, Thien},
  booktitle={Findings of the Association for Computational Linguistics: ACL 2024},
  month={aug},
  address={Bangkok, Thailand},
  publisher={Association for Computational Linguistics},
  pages={1038--1047},
  year={2024},
  url={https://aclanthology.org/2024.findings-acl.61/},
  doi={10.18653/v1/2024.findings-acl.61}
}

@inproceedings{yi2024position,
  title={Position: building guardrails for large language models requires systematic design},
  author={Yi, DONG and Mu, Ronghui and Jin, Gaojie and Qi, Yi and Hu, Jinwei and Zhao, Xingyu and Meng, Jie and Ruan, Wenjie and Huang, Xiaowei},
  booktitle={Forty-first International Conference on Machine Learning},
  pages={11375--11394},
  year={2024},
  url={https://proceedings.mlr.press/v235/dong24c.html}
}

@inproceedings{verma2025multiguard,
  title={MULTIGUARD: An Efficient Approach for AI Safety Moderation Across Languages and Modalities},
  author={Verma, Sahil and Hines, Keegan and Bilmes, Jeff and Siska, Charlotte and Zettlemoyer, Luke and Gonen, Hila and Singh, Chandan},
  booktitle={Proceedings of the 2025 Conference on Empirical Methods in Natural Language Processing},
  month={nov},
  address={Suzhou, China},
  publisher={Association for Computational Linguistics},
  pages={16173--16187},
  year={2025},
  url={https://aclanthology.org/2025.emnlp-main.819/},
  doi={10.18653/v1/2025.emnlp-main.819}
}

@article{kumar2025polyguard,
  title={Polyguard: A multilingual safety moderation tool for 17 languages},
  author={Kumar, Priyanshu and Jain, Devansh and Yerukola, Akhila and Jiang, Liwei and Beniwal, Himanshu and Hartvigsen, Thomas and Sap, Maarten},
  journal={arXiv preprint arXiv:2504.04377},
  year={2025},
  eprint={2504.04377},
  archivePrefix={arXiv},
  primaryClass={cs.CL},
  url={https://arxiv.org/abs/2504.04377},
  doi={10.48550/arXiv.2504.04377}
}

@article{inan2023llama,
  title={Llama guard: Llm-based input-output safeguard for human-ai conversations},
  author={Inan, Hakan and Upasani, Kartikeya and Chi, Jianfeng and Rungta, Rashi and Iyer, Krithika and Mao, Yuning and Tontchev, Michael and Hu, Qing and Fuller, Brian and Testuggine, Davide},
  journal={arXiv preprint arXiv:2312.06674},
  year={2023},
  eprint={2312.06674},
  archivePrefix={arXiv},
  primaryClass={cs.CL},
  url={https://arxiv.org/abs/2312.06674},
  doi={10.48550/arXiv.2312.06674}
}

@misc{ghosh2024aegis,
      title={AEGIS: Online Adaptive AI Content Safety Moderation with Ensemble of LLM Experts}, 
      author={Shaona Ghosh and Prasoon Varshney and Erick Galinkin and Christopher Parisien},
      year={2024},
      eprint={2404.05993},
      archivePrefix={arXiv},
      primaryClass={cs.LG},
      url={https://arxiv.org/abs/2404.05993}, 
}

@article{han2024wildguard,
  title={Wildguard: Open one-stop moderation tools for safety risks, jailbreaks, and refusals of llms},
  author={Han, Seungju and Rao, Kavel and Ettinger, Allyson and Jiang, Liwei and Lin, Bill Yuchen and Lambert, Nathan and Choi, Yejin and Dziri, Nouha},
  journal={Advances in Neural Information Processing Systems},
  volume={37},
  pages={8093--8131},
  year={2024},
  url={https://proceedings.neurips.cc/paper_files/paper/2024/hash/0f69b4b96a46f284b726fbd70f74fb3b-Abstract-Datasets_and_Benchmarks_Track.html}
}

@article{bansal2025crest,
  title={CREST: Universal Safety Guardrails Through Cluster-Guided Cross-Lingual Transfer},
  author={Bansal, Lavish and Mishra, Naman},
  journal={arXiv preprint arXiv:2512.02711},
  year={2025},
  eprint={2512.02711},
  archivePrefix={arXiv},
  primaryClass={cs.CL},
  url={https://arxiv.org/abs/2512.02711},
  doi={10.48550/arXiv.2512.02711},
  note={Accepted at LREC 2026}
}

@inproceedings{yang-etal-2025-mrguard,
    title = "{M}r{G}uard: A Multilingual Reasoning Guardrail for Universal {LLM} Safety",
    author = "Yang, Yahan  and
      Dan, Soham  and
      Li, Shuo  and
      Roth, Dan  and
      Lee, Insup",
    editor = "Christodoulopoulos, Christos  and
      Chakraborty, Tanmoy  and
      Rose, Carolyn  and
      Peng, Violet",
    booktitle = "Proceedings of the 2025 Conference on Empirical Methods in Natural Language Processing",
    month = nov,
    year = "2025",
    address = "Suzhou, China",
    publisher = "Association for Computational Linguistics",
    url = "https://aclanthology.org/2025.emnlp-main.1392/",
    doi = "10.18653/v1/2025.emnlp-main.1392",
    pages = "27377--27396",
    ISBN = "979-8-89176-332-6"
}

@inproceedings{upadhayay-behzadan-2025-tongue,
    title = "Tongue-Tied: Breaking {LLM}s Safety Through New Language Learning",
    author = "Upadhayay, Bibek  and
      Behzadan, Vahid",
    editor = "Winata, Genta Indra  and
      Kar, Sudipta  and
      Zhukova, Marina  and
      Solorio, Thamar  and
      Ai, Xi  and
      Hamed, Injy  and
      Ihsani, Mahardika Krisna Krisna  and
      Wijaya, Derry Tanti  and
      Kuwanto, Garry",
    booktitle = "Proceedings of the 7th Workshop on Computational Approaches to Linguistic Code-Switching",
    month = may,
    year = "2025",
    address = "Albuquerque, New Mexico, USA",
    publisher = "Association for Computational Linguistics",
    url = "https://aclanthology.org/2025.calcs-1.5/",
    doi = "10.18653/v1/2025.calcs-1.5",
    pages = "32--47",
    ISBN = "979-8-89176-053-0"
}

@inproceedings{al2024jailbreaking,
  title={Jailbreaking llms with arabic transliteration and arabizi},
  author={Al Ghanim, Mansour and Almohaimeed, Saleh and Zheng, Mengxin and Solihin, Yan and Lou, Qian},
  booktitle={Proceedings of the 2024 conference on empirical methods in natural language processing},
  month={nov},
  address={Miami, Florida, USA},
  publisher={Association for Computational Linguistics},
  pages={18584--18600},
  year={2024},
  url={https://aclanthology.org/2024.emnlp-main.1034/},
  doi={10.18653/v1/2024.emnlp-main.1034}
}

@inproceedings{yoo2025code,
  title={Code-switching red-teaming: Llm evaluation for safety and multilingual understanding},
  author={Yoo, Haneul and Yang, Yongjin and Lee, Hwaran},
  booktitle={Proceedings of the 63rd Annual Meeting of the Association for Computational Linguistics (Volume 1: Long Papers)},
  month={jul},
  address={Vienna, Austria},
  publisher={Association for Computational Linguistics},
  pages={13392--13413},
  year={2025},
  url={https://aclanthology.org/2025.acl-long.657/},
  doi={10.18653/v1/2025.acl-long.657}
}

@inproceedings{zhang2025english,
  title={English as Defense Proxy: Mitigating Multilingual Jailbreak via Eliciting English Safety Knowledge},
  author={Zhang, Zekai and Guo, Yiduo and Lin, Jiuheng and Quan, Shanghaoran and Zhang, Huishuai and Zhao, Dongyan},
  booktitle={Findings of the Association for Computational Linguistics: EMNLP 2025},
  month={nov},
  address={Suzhou, China},
  publisher={Association for Computational Linguistics},
  pages={1185--1196},
  year={2025},
  url={https://aclanthology.org/2025.findings-emnlp.62/},
  doi={10.18653/v1/2025.findings-emnlp.62}
}

@inproceedings{poppi2025towards,
  title={Towards understanding the fragility of multilingual llms against fine-tuning attacks},
  author={Poppi, Samuele and Yong, Zheng-Xin and He, Yifei and Chern, Bobbie and Zhao, Han and Yang, Aobo and Chi, Jianfeng},
  booktitle={Findings of the Association for Computational Linguistics: NAACL 2025},
  month={apr},
  address={Albuquerque, New Mexico},
  publisher={Association for Computational Linguistics},
  pages={2358--2372},
  year={2025},
  url={https://aclanthology.org/2025.findings-naacl.126/},
  doi={10.18653/v1/2025.findings-naacl.126}
}

@inproceedings{dementieva-etal-2023-exploring,
    title = "Exploring Methods for Cross-lingual Text Style Transfer: The Case of Text Detoxification",
    author = "Dementieva, Daryna  and
      Moskovskiy, Daniil  and
      Dale, David  and
      Panchenko, Alexander",
    editor = "Park, Jong C.  and
      Arase, Yuki  and
      Hu, Baotian  and
      Lu, Wei  and
      Wijaya, Derry  and
      Purwarianti, Ayu  and
      Krisnadhi, Adila Alfa",
    booktitle = "Proceedings of the 13th International Joint Conference on Natural Language Processing and the 3rd Conference of the Asia-Pacific Chapter of the Association for Computational Linguistics (Volume 1: Long Papers)",
    month = nov,
    year = "2023",
    address = "Nusa Dua, Bali",
    publisher = "Association for Computational Linguistics",
    url = "https://aclanthology.org/2023.ijcnlp-main.70/",
    doi = "10.18653/v1/2023.ijcnlp-main.70",
    pages = "1083--1101"
}

@inproceedings{multiparadetox,
    title = "{M}ulti{P}ara{D}etox: Extending Text Detoxification with Parallel Data to New Languages",
    author = "Dementieva, Daryna  and
      Babakov, Nikolay  and
      Panchenko, Alexander",
    editor = "Duh, Kevin  and
      Gomez, Helena  and
      Bethard, Steven",
    booktitle = "Proceedings of the 2024 Conference of the North American Chapter of the Association for Computational Linguistics: Human Language Technologies (Volume 2: Short Papers)",
    month = jun,
    year = "2024",
    address = "Mexico City, Mexico",
    publisher = "Association for Computational Linguistics",
    url = "https://aclanthology.org/2024.naacl-short.12/",
    doi = "10.18653/v1/2024.naacl-short.12",
    pages = "124--140"
}

@inproceedings{sellam2020bleurt,
  title={BLEURT: Learning Robust Metrics for Text Generation},
  author={Sellam, Thibault and Das, Dipanjan and Parikh, Ankur},
  booktitle={Proceedings of the 58th Annual Meeting of the Association for Computational Linguistics},
  month={jul},
  address={Online},
  publisher={Association for Computational Linguistics},
  pages={7881--7892},
  year={2020},
  url={https://aclanthology.org/2020.acl-main.704/},
  doi={10.18653/v1/2020.acl-main.704}
}

@inproceedings{sap-etal-2019-risk,
    title = "The Risk of Racial Bias in Hate Speech Detection",
    author = "Sap, Maarten  and
      Card, Dallas  and
      Gabriel, Saadia  and
      Choi, Yejin  and
      Smith, Noah A.",
    editor = "Korhonen, Anna  and
      Traum, David  and
      M{\`a}rquez, Llu{\'i}s",
    booktitle = "Proceedings of the 57th Annual Meeting of the Association for Computational Linguistics",
    month = jul,
    year = "2019",
    address = "Florence, Italy",
    publisher = "Association for Computational Linguistics",
    url = "https://aclanthology.org/P19-1163/",
    doi = "10.18653/v1/P19-1163",
    pages = "1668--1678"
}

@misc{turner2024steeringlanguagemodelsactivation,
      title={Steering Language Models With Activation Engineering}, 
      author={Alexander Matt Turner and Lisa Thiergart and Gavin Leech and David Udell and Juan J. Vazquez and Ulisse Mini and Monte MacDiarmid},
      year={2024},
      eprint={2308.10248},
      archivePrefix={arXiv},
      primaryClass={cs.CL},
      url={https://arxiv.org/abs/2308.10248}, 
}

@inproceedings{meng2022locating,
  title={Locating and Editing Factual Associations in GPT},
  author={Meng, Kevin and Bau, David and Andonian, Alex and Belinkov, Yonatan},
  booktitle={Advances in Neural Information Processing Systems},
  volume={35},
  pages={17359--17372},
  year={2022},
  url={https://proceedings.neurips.cc/paper_files/paper/2022/hash/6f1d43d5a82a37e89b0665b33bf3a182-Abstract-Conference.html}
}

@inproceedings{ouyang2022training,
  title={Training Language Models to Follow Instructions with Human Feedback},
  author={Ouyang, Long and Wu, Jeffrey and Jiang, Xu and Almeida, Diogo and Wainwright, Carroll and Mishkin, Pamela and Zhang, Chong and Agarwal, Sandhini and Slama, Katarina and Ray, Alex and Schulman, John and Hilton, Jacob and Kelton, Fraser and Miller, Luke and Simens, Maddie and Askell, Amanda and Welinder, Peter and Christiano, Paul F. and Leike, Jan and Lowe, Ryan},
  booktitle={Advances in Neural Information Processing Systems},
  volume={35},
  pages={27730--27744},
  year={2022},
  url={https://proceedings.neurips.cc/paper_files/paper/2022/hash/b1efde53be364a73914f58805a001731-Abstract-Conference.html}
}

@article{zhuo2023red,
  title={Red teaming chatgpt via jailbreaking: Bias, robustness, reliability and toxicity},
  author={Zhuo, Terry Yue and Huang, Yujin and Chen, Chunyang and Xing, Zhenchang},
  journal={arXiv preprint arXiv:2301.12867},
  year={2023},
  eprint={2301.12867},
  archivePrefix={arXiv},
  primaryClass={cs.CL},
  url={https://arxiv.org/abs/2301.12867},
  doi={10.48550/arXiv.2301.12867}
}

@inproceedings{cecchini2024holistic,
  title={Holistic evaluation of large language models: Assessing robustness, accuracy, and toxicity for real-world applications},
  author={Cecchini, David and Nazir, Arshaan and Chakravarthy, Kalyan and Kocaman, Veysel},
  booktitle={Proceedings of the 4th Workshop on Trustworthy Natural Language Processing (TrustNLP 2024)},
  month={jun},
  address={Mexico City, Mexico},
  publisher={Association for Computational Linguistics},
  pages={109--117},
  year={2024},
  url={https://aclanthology.org/2024.trustnlp-1.11/},
  doi={10.18653/v1/2024.trustnlp-1.11}
}

@article{adragna2020fairness,
  title={Fairness and robustness in invariant learning: A case study in toxicity classification},
  author={Adragna, Robert and Creager, Elliot and Madras, David and Zemel, Richard},
  journal={arXiv preprint arXiv:2011.06485},
  year={2020},
  eprint={2011.06485},
  archivePrefix={arXiv},
  primaryClass={cs.LG},
  url={https://arxiv.org/abs/2011.06485},
  doi={10.48550/arXiv.2011.06485}
}

@inproceedings{costa2023toxicity,
  title={Toxicity in multilingual machine translation at scale},
  author={Costa-Juss{\`a}, Marta R. and Smith, Eric and Ropers, Christophe and Licht, Daniel and Maillard, Jean and Ferrando, Javier and Escolano, Carlos},
  booktitle={Findings of the Association for Computational Linguistics: EMNLP 2023},
  month={dec},
  address={Singapore},
  publisher={Association for Computational Linguistics},
  pages={9570--9586},
  year={2023},
  url={https://aclanthology.org/2023.findings-emnlp.642/},
  doi={10.18653/v1/2023.findings-emnlp.642}
}

@article{huang2025content,
  title={Content moderation by LLM: from accuracy to legitimacy},
  author={Huang, Tao},
  journal={Artificial Intelligence Review},
  volume={58},
  number={10},
  pages={320},
  year={2025},
  publisher={Springer},
  url={https://link.springer.com/article/10.1007/s10462-025-11328-1},
  doi={10.1007/s10462-025-11328-1}
}

@inproceedings{dammu2024they,
  title={“They are uncultured”: Unveiling Covert Harms and Social Threats in LLM Generated Conversations},
  author={Dammu, Preetam Prabhu Srikar and Jung, Hayoung and Singh, Anjali and Choudhury, Monojit and Mitra, Tanu},
  booktitle={Proceedings of the 2024 Conference on Empirical Methods in Natural Language Processing},
  month={nov},
  address={Miami, Florida, USA},
  publisher={Association for Computational Linguistics},
  pages={20339--20369},
  year={2024},
  url={https://aclanthology.org/2024.emnlp-main.1134/},
  doi={10.18653/v1/2024.emnlp-main.1134}
}

@inproceedings{vongpradit2024safecultural,
  title={SafeCultural: A Dataset for Evaluating Safety and Cultural Sensitivity in Large Language Models},
  author={Vongpradit, Pawat and Imsombut, Aurawan and Kongyoung, Sarawoot and Damrongrat, Chaianun and Phaholphinyo, Sitthaa and Tanawong, Tanik},
  booktitle={2024 8th International Conference on Information Technology (InCIT)},
  pages={740--745},
  year={2024},
  organization={IEEE},
  doi={10.1109/InCIT63192.2024.10810548}
}

@inproceedings{wen2023unveiling,
  title={Unveiling the Implicit Toxicity in Large Language Models},
  author={Wen, Jiaxin and Ke, Pei and Sun, Hao and Zhang, Zhexin and Li, Chengfei and Bai, Jinfeng and Huang, Minlie},
  booktitle={Proceedings of the 2023 Conference on Empirical Methods in Natural Language Processing},
  month={dec},
  address={Singapore},
  publisher={Association for Computational Linguistics},
  pages={1322--1338},
  year={2023},
  url={https://aclanthology.org/2023.emnlp-main.84/},
  doi={10.18653/v1/2023.emnlp-main.84}
}

@misc{beniwal2025breakingmbadsupervisedfinetuning,
      title={Breaking mBad! Supervised Fine-tuning for Cross-Lingual Detoxification}, 
      author={Himanshu Beniwal and Youngwoo Kim and Maarten Sap and Soham Dan and Thomas Hartvigsen},
      year={2025},
      eprint={2505.16722},
      archivePrefix={arXiv},
      primaryClass={cs.CL},
      url={https://arxiv.org/abs/2505.16722}, 
}

@inproceedings{kim2025decoding,
  title={Decoding the Rule Book: Extracting Hidden Moderation Criteria from Reddit Communities},
  author={Kim, Youngwoo and Beniwal, Himanshu and Johnson, Steven L and Hartvigsen, Thomas},
  booktitle={Proceedings of the 2025 Conference on Empirical Methods in Natural Language Processing},
  month={nov},
  address={Suzhou, China},
  publisher={Association for Computational Linguistics},
  pages={20487--20498},
  year={2025},
  url={https://aclanthology.org/2025.emnlp-main.1034/},
  doi={10.18653/v1/2025.emnlp-main.1034}
}

@misc{bai2022constitutionalai,
  title={Constitutional AI: Harmlessness from AI Feedback},
  author={Bai, Yuntao and Kadavath, Saurav and Kundu, Sandipan and Askell, Amanda and Kernion, Jackson and Jones, Andy and Chen, Anna and Goldie, Anna and Mirhoseini, Azalia and McKinnon, Cameron and Chen, Carol and Olsson, Catherine and Olah, Christopher and Hernandez, Danny and Drain, Dawn and Ganguli, Deep and Li, Dustin and Tran-Johnson, Eli and Perez, Ethan and Kerr, Jamie and Mueller, Jared and Ladish, Jeffrey and Landau, Joshua and Ndousse, Kamal and Lukosiute, Kamile and Lovitt, Liane and Sellitto, Michael and Elhage, Nelson and Schiefer, Nicholas and Mercado, Noem{\'i} and DasSarma, Nova and Lasenby, Robert and Larson, Robin and Ringer, Sam and Johnston, Scott and Kravec, Shauna and El Showk, Sheer and Fort, Stanislav and Lanham, Tamera and Telleen-Lawton, Timothy and Conerly, Tom and Henighan, Tom and Hume, Tristan and Bowman, Samuel R. and Hatfield-Dodds, Zac and Mann, Ben and Amodei, Dario and Joseph, Nicholas and McCandlish, Sam and Brown, Tom and Kaplan, Jared},
  year={2022},
  eprint={2212.08073},
  archivePrefix={arXiv},
  primaryClass={cs.CL},
  url={https://arxiv.org/abs/2212.08073}
}

@article{Kreutzer_2022,
   title={Quality at a Glance: An Audit of Web-Crawled Multilingual Datasets},
   volume={10},
   ISSN={2307-387X},
   url={http://dx.doi.org/10.1162/tacl_a_00447},
   DOI={10.1162/tacl_a_00447},
   journal={Transactions of the Association for Computational Linguistics},
   publisher={MIT Press},
   author={Kreutzer, Julia and Caswell, Isaac and Wang, Lisa and Wahab, Ahsan and van Esch, Daan and Ulzii-Orshikh, Nasanbayar and Tapo, Allahsera and Subramani, Nishant and Sokolov, Artem and Sikasote, Claytone and Setyawan, Monang and Sarin, Supheakmungkol and Samb, Sokhar and Sagot, Benoît and Rivera, Clara and Rios, Annette and Papadimitriou, Isabel and Osei, Salomey and Suarez, Pedro Ortiz and Orife, Iroro and Ogueji, Kelechi and Rubungo, Andre Niyongabo and Nguyen, Toan Q. and Müller, Mathias and Müller, André and Muhammad, Shamsuddeen Hassan and Muhammad, Nanda and Mnyakeni, Ayanda and Mirzakhalov, Jamshidbek and Matangira, Tapiwanashe and Leong, Colin and Lawson, Nze and Kudugunta, Sneha and Jernite, Yacine and Jenny, Mathias and Firat, Orhan and Dossou, Bonaventure F. P. and Dlamini, Sakhile and de Silva, Nisansa and Çabuk Ballı, Sakine and Biderman, Stella and Battisti, Alessia and Baruwa, Ahmed and Bapna, Ankur and Baljekar, Pallavi and Azime, Israel Abebe and Awokoya, Ayodele and Ataman, Duygu and Ahia, Orevaoghene and Ahia, Oghenefego and Agrawal, Sweta and Adeyemi, Mofetoluwa},
   year={2022},
   pages={50–72} }

@misc{lu2025alignmentsafetylargelanguage,
      title={Alignment and Safety in Large Language Models: Safety Mechanisms, Training Paradigms, and Emerging Challenges}, 
      author={Haoran Lu and Luyang Fang and Ruidong Zhang and Xinliang Li and Jiazhang Cai and Huimin Cheng and Lin Tang and Ziyu Liu and Zeliang Sun and Tao Wang and Yingchuan Zhang and Arif Hassan Zidan and Jinwen Xu and Jincheng Yu and Meizhi Yu and Hanqi Jiang and Xilin Gong and Weidi Luo and Bolun Sun and Yongkai Chen and Terry Ma and Shushan Wu and Yifan Zhou and Junhao Chen and Haotian Xiang and Jing Zhang and Afrar Jahin and Wei Ruan and Ke Deng and Yi Pan and Peilong Wang and Jiahui Li and Zhengliang Liu and Lu Zhang and Lin Zhao and Wei Liu and Dajiang Zhu and Xin Xing and Fei Dou and Wei Zhang and Chao Huang and Rongjie Liu and Mengrui Zhang and Yiwen Liu and Xiaoxiao Sun and Qin Lu and Zhen Xiang and Wenxuan Zhong and Tianming Liu and Ping Ma},
      year={2025},
      eprint={2507.19672},
      archivePrefix={arXiv},
      primaryClass={cs.AI},
      url={https://arxiv.org/abs/2507.19672}, 
}

@misc{bensalem2024toxiclanguagedetectionsystematic,
      title={Toxic language detection: a systematic review of Arabic datasets}, 
      author={Imene Bensalem and Paolo Rosso and Hanane Zitouni},
      year={2024},
      eprint={2312.07228},
      archivePrefix={arXiv},
      primaryClass={cs.CL},
      url={https://arxiv.org/abs/2312.07228}, 
}

@inproceedings{paradetox,
    title = "{P}ara{D}etox: Detoxification with Parallel Data",
    author = "Logacheva, Varvara  and
      Dementieva, Daryna  and
      Ustyantsev, Sergey  and
      Moskovskiy, Daniil  and
      Dale, David  and
      Krotova, Irina  and
      Semenov, Nikita  and
      Panchenko, Alexander",
    editor = "Muresan, Smaranda  and
      Nakov, Preslav  and
      Villavicencio, Aline",
    booktitle = "Proceedings of the 60th Annual Meeting of the Association for Computational Linguistics (Volume 1: Long Papers)",
    month = may,
    year = "2022",
    address = "Dublin, Ireland",
    publisher = "Association for Computational Linguistics",
    url = "https://aclanthology.org/2022.acl-long.469/",
    doi = "10.18653/v1/2022.acl-long.469",
    pages = "6804--6818"
}

@inproceedings{safedit,
    title = "Detoxifying Large Language Models via Knowledge Editing",
    author = "Wang, Mengru  and
      Zhang, Ningyu  and
      Xu, Ziwen  and
      Xi, Zekun  and
      Deng, Shumin  and
      Yao, Yunzhi  and
      Zhang, Qishen  and
      Yang, Linyi  and
      Wang, Jindong  and
      Chen, Huajun",
    editor = "Ku, Lun-Wei  and
      Martins, Andre  and
      Srikumar, Vivek",
    booktitle = "Proceedings of the 62nd Annual Meeting of the Association for Computational Linguistics (Volume 1: Long Papers)",
    month = aug,
    year = "2024",
    address = "Bangkok, Thailand",
    publisher = "Association for Computational Linguistics",
    url = "https://aclanthology.org/2024.acl-long.171/",
    doi = "10.18653/v1/2024.acl-long.171",
    pages = "3093--3118"
}

@inproceedings{gehman-etal-2020-realtoxicityprompts,
  title = {RealToxicityPrompts: Evaluating Neural Toxic Degeneration in Language Models},
  author = {Gehman, Samuel and Gururangan, Suchin and Sap, Maarten and Choi, Yejin and Smith, Noah A.},
  booktitle = {Findings of the Association for Computational Linguistics: EMNLP 2020},
  year = {2020},
  pages = {3356--3369},
  doi = {10.18653/v1/2020.findings-emnlp.301},
  url = {https://aclanthology.org/2020.findings-emnlp.301}
}

@inproceedings{koh-etal-2024-llms,
  title = {Can LLMs Recognize Toxicity? A Structured Investigation Framework and Toxicity Metric},
  author = {Koh, Hyukhun and Kim, Dohyung and Lee, Minwoo and Jung, Kyomin},
  booktitle = {Findings of the Association for Computational Linguistics: EMNLP 2024},
  month = nov,
  address = {Miami, Florida, USA},
  publisher = {Association for Computational Linguistics},
  year = {2024},
  pages = {6092--6114},
  doi = {10.18653/v1/2024.findings-emnlp.353},
  url = {https://aclanthology.org/2024.findings-emnlp.353}
}

@inproceedings{deshpande-etal-2023-chatgpt,
  title = {Toxicity in ChatGPT: Analyzing Persona-assigned Language Models},
  author = {Deshpande, Ameet and Murahari, Vishvak and Rajpurohit, Tanmay and Kalyan, Ashwin and Narasimhan, Karthik},
  booktitle = {Findings of the Association for Computational Linguistics: EMNLP 2023},
  month = dec,
  address = {Singapore},
  publisher = {Association for Computational Linguistics},
  year = {2023},
  pages = {1236--1270},
  url = {https://aclanthology.org/2023.findings-emnlp.88/},
  doi = {10.18653/v1/2023.findings-emnlp.88}
}

@misc{beniwal2025unityaiguardpioneeringtoxicitydetection,
      title={UNITYAI-GUARD: Pioneering Toxicity Detection Across Low-Resource Indian Languages}, 
      author={Himanshu Beniwal and Reddybathuni Venkat and Rohit Kumar and Birudugadda Srivibhav and Daksh Jain and Pavan Doddi and Eshwar Dhande and Adithya Ananth and Kuldeep and Mayank Singh},
      year={2025},
      eprint={2503.23088},
      archivePrefix={arXiv},
      primaryClass={cs.CL},
      url={https://arxiv.org/abs/2503.23088}, 
}

@article{jain-etal-2024-polygloToxicityprompts,
    author    = {Devansh Jain and Priyanshu Kumar and Samuel Gehman and Xuhui Zhou and Thomas Hartvigsen and Maarten Sap},
    title     = {PolygloToxicityPrompts: Multilingual Evaluation of Neural Toxic Degeneration in Large Language Models},
    journal   = {arXiv preprint},
    volume    = {arXiv:2405.09373},
    year      = {2024},
    url       = {https://arxiv.org/abs/2405.09373},
    note      = {May 2024} 
}

@article{li-yong-bach-2024-preference,
    author    = {Xiaochen Li and Zheng{-}Xin Yong and Stephen H. Bach},
    title     = {Preference Tuning for Toxicity Mitigation Generalizes Across Languages},
    journal   = {arXiv preprint},
    volume    = {arXiv:2406.16235},
    year      = {2024},
    url       = {https://arxiv.org/abs/2406.16235},
    note      = {June 2024} 
}

@inproceedings{jaggi-etal-2024-accurate,
    title = "Accurate and Data-Efficient Toxicity Prediction when Annotators Disagree",
    author = "Jaggi, Harbani  and
      Coimbatore Murali, Kashyap  and
      Fleisig, Eve  and
      Biyik, Erdem",
    editor = "Al-Onaizan, Yaser  and
      Bansal, Mohit  and
      Chen, Yun-Nung",
    booktitle = "Proceedings of the 2024 Conference on Empirical Methods in Natural Language Processing",
    month = nov,
    year = "2024",
    address = "Miami, Florida, USA",
    publisher = "Association for Computational Linguistics",
    url = "https://aclanthology.org/2024.emnlp-main.1221/",
    doi = "10.18653/v1/2024.emnlp-main.1221",
    pages = "21910--21917"
}

@inproceedings{pascual-etal-2021-plug-play,
    title = "A Plug-and-Play Method for Controlled Text Generation",
    author = "Pascual, Damian  and
      Egressy, Beni  and
      Meister, Clara  and
      Cotterell, Ryan  and
      Wattenhofer, Roger",
    editor = "Moens, Marie-Francine  and
      Huang, Xuanjing  and
      Specia, Lucia  and
      Yih, Scott Wen-tau",
    booktitle = "Findings of the Association for Computational Linguistics: EMNLP 2021",
    month = nov,
    year = "2021",
    address = "Punta Cana, Dominican Republic",
    publisher = "Association for Computational Linguistics",
    url = "https://aclanthology.org/2021.findings-emnlp.334/",
    doi = "10.18653/v1/2021.findings-emnlp.334",
    pages = "3973--3997"
}

@inproceedings{liu-etal-2021-dexperts,
    author    = {Alisa Liu and Maarten Sap and Ximing Lu and Swabha Swayamdipta and Chandra Bhagavatula and Noah A. Smith and Yejin Choi},
    title     = {DExperts: Decoding-Time Controlled Text Generation with Experts and Anti-Experts},
    booktitle = {Proceedings of the 59th Annual Meeting of the Association for Computational Linguistics (ACL 2021)},
    pages     = {1990--2001},
    year      = {2021},
    address   = {Online},
    publisher = {Association for Computational Linguistics},
    url       = {https://aclanthology.org/2021.acl-long.522/},
    note      = {ACL 2021} 
}

@inproceedings{xu2021detoxifying,
  title={Detoxifying Language Models Risks Marginalizing Minority Voices},
  author={Xu, Albert and Pathak, Eshaan and Wallace, Eric and Gururangan, Suchin and Sap, Maarten and Klein, Dan},
  booktitle={Proceedings of the 2021 Conference of the North American Chapter of the Association for Computational Linguistics: Human Language Technologies},
  month={jun},
  address={Online},
  publisher={Association for Computational Linguistics},
  pages={2390--2397},
  year={2021},
  url={https://aclanthology.org/2021.naacl-main.190/},
  doi={10.18653/v1/2021.naacl-main.190}
}

@inproceedings{pozzobon-etal-2023-goodtriever,
    author    = {Luiza Pozzobon and Beyza Ermis and Patrick Lewis and Sara Hooker},
    title     = {Goodtriever: Adaptive Toxicity Mitigation with Retrieval-augmented Models},
    booktitle = {Findings of the Association for Computational Linguistics: EMNLP 2023},
    pages     = {5108--5125},
    month     = {December},
    year      = {2023},
    address   = {Singapore},
    publisher = {Association for Computational Linguistics},
    url       = {https://aclanthology.org/2023.findings-emnlp.339/},
    doi       = {10.18653/v1/2023.findings-emnlp.339},
    note      = {EMNLP Findings 2023} 
}

@inproceedings{atwell2022appdia,
  title={APPDIA: A Discourse-aware Transformer-based Style Transfer Model for Offensive Social Media Conversations},
  author={Atwell, Katherine and Hassan, Sabit and Alikhani, Malihe},
  booktitle={Proceedings of the 29th International Conference on Computational Linguistics},
  month={oct},
  address={Gyeongju, Republic of Korea},
  publisher={International Committee on Computational Linguistics},
  pages={6063--6074},
  year={2022},
  url={https://aclanthology.org/2022.coling-1.530/}
}

@inproceedings{leong2023self,
  title={Self-Detoxifying Language Models via Toxification Reversal},
  author={Leong, Chak Tou and Cheng, Yi and Wang, Jiashuo and Wang, Jian and Li, Wenjie},
  booktitle={Proceedings of the 2023 Conference on Empirical Methods in Natural Language Processing},
  pages={4433--4449},
  year={2023},
  address={Singapore},
  publisher={Association for Computational Linguistics},
  url={https://aclanthology.org/2023.emnlp-main.269/},
  doi={10.18653/v1/2023.emnlp-main.269}
}

@article{moskovskiy2025synthdetoxm,
  title={SynthDetoxM: Modern LLMs are Few-Shot Parallel Detoxification Data Annotators},
  author={Moskovskiy, Daniil and Sushko, Nikita and Pletenev, Sergey and Tutubalina, Elena and Panchenko, Alexander},
  journal={arXiv preprint arXiv:2502.06394},
  year={2025},
  eprint={2502.06394},
  archivePrefix={arXiv},
  primaryClass={cs.CL},
  url={https://arxiv.org/abs/2502.06394},
  doi={10.48550/arXiv.2502.06394}
}

@inproceedings{neplenbroek2025cross,
  title={Cross-lingual transfer of debiasing and detoxification in multilingual llms: An extensive investigation},
  author={Neplenbroek, Vera and Bisazza, Arianna and Fern{\'a}ndez, Raquel},
  booktitle={Findings of the Association for Computational Linguistics: ACL 2025},
  month={jul},
  address={Vienna, Austria},
  publisher={Association for Computational Linguistics},
  pages={2805--2830},
  year={2025},
  url={https://aclanthology.org/2025.findings-acl.145/},
  doi={10.18653/v1/2025.findings-acl.145}
}

@misc{jigsaw-multilingual-toxic-comment-classification,
    author = {Ian Kivlichan and Jeffrey Sorensen and Julia Elliott and Lucy Vasserman and Martin Görner and Phil Culliton},
    title = {Jigsaw Multilingual Toxic Comment Classification},
    year = {2020},
    howpublished = {Kaggle competition},
    url = {https://www.kaggle.com/c/jigsaw-multilingual-toxic-comment-classification}
}

@inproceedings{zampieri2019semeval,
  title={SemEval-2019 Task 6: Identifying and Categorizing Offensive Language in Social Media (OffensEval)},
  author={Zampieri, Marcos and Malmasi, Shervin and Nakov, Preslav and Rosenthal, Sara and Farra, Noura and Kumar, Ritesh},
  booktitle={Proceedings of the 13th International Workshop on Semantic Evaluation},
  month={jun},
  address={Minneapolis, Minnesota, USA},
  publisher={Association for Computational Linguistics},
  pages={75--86},
  year={2019},
  url={https://aclanthology.org/S19-2010/},
  doi={10.18653/v1/S19-2010}
}

@inproceedings{pavlopoulos-etal-2021-semeval,
    title = "{S}em{E}val-2021 Task 5: Toxic Spans Detection",
    author = "Pavlopoulos, John  and
      Sorensen, Jeffrey  and
      Laugier, L{\'e}o  and
      Androutsopoulos, Ion",
    editor = "Palmer, Alexis  and
      Schneider, Nathan  and
      Schluter, Natalie  and
      Emerson, Guy  and
      Herbelot, Aurelie  and
      Zhu, Xiaodan",
    booktitle = "Proceedings of the 15th International Workshop on Semantic Evaluation (SemEval-2021)",
    month = aug,
    year = "2021",
    address = "Online",
    publisher = "Association for Computational Linguistics",
    url = "https://aclanthology.org/2021.semeval-1.6/",
    doi = "10.18653/v1/2021.semeval-1.6",
    pages = "59--69"
}

@inproceedings{rottger2021hatecheck,
  title={HateCheck: Functional Tests for Hate Speech Detection Models},
  author={R{\"o}ttger, Paul and Vidgen, Bertie and Nguyen, Dong and Waseem, Zeerak and Margetts, Helen and Pierrehumbert, Janet},
  booktitle={Proceedings of the 59th Annual Meeting of the Association for Computational Linguistics and the 11th International Joint Conference on Natural Language Processing (Volume 1: Long Papers)},
  month={aug},
  address={Online},
  publisher={Association for Computational Linguistics},
  pages={41--58},
  year={2021},
  url={https://aclanthology.org/2021.acl-long.4/},
  doi={10.18653/v1/2021.acl-long.4}
}

@inproceedings{rottger2022multilingual,
  title={Multilingual HateCheck: Functional Tests for Multilingual Hate Speech Detection Models},
  author={R{\"o}ttger, Paul and Seelawi, Haitham and Nozza, Debora and Talat, Zeerak and Vidgen, Bertie},
  booktitle={Proceedings of the Sixth Workshop on Online Abuse and Harms (WOAH)},
  month={jul},
  address={Seattle, Washington (Hybrid)},
  publisher={Association for Computational Linguistics},
  pages={154--169},
  year={2022},
  url={https://aclanthology.org/2022.woah-1.15/},
  doi={10.18653/v1/2022.woah-1.15}
}

@inproceedings{basile2019semeval,
  title={Semeval-2019 task 5: Multilingual detection of hate speech against immigrants and women in twitter},
  author={Basile, Valerio and Bosco, Cristina and Fersini, Elisabetta and Nozza, Debora and Patti, Viviana and Pardo, Francisco Manuel Rangel and Rosso, Paolo and Sanguinetti, Manuela},
  booktitle={Proceedings of the 13th international workshop on semantic evaluation},
  month={jun},
  address={Minneapolis, Minnesota, USA},
  publisher={Association for Computational Linguistics},
  pages={54--63},
  year={2019},
  url={https://aclanthology.org/S19-2007/},
  doi={10.18653/v1/S19-2007}
}

@inproceedings{mandl2019overview,
  title={Overview of the hasoc track at fire 2019: Hate speech and offensive content identification in indo-european languages},
  author={Mandl, Thomas and Modha, Sandip and Majumder, Prasenjit and Patel, Daksh and Dave, Mohana and Mandlia, Chintak and Patel, Aditya},
  booktitle={Proceedings of the 11th annual meeting of the Forum for Information Retrieval Evaluation},
  pages={14--17},
  year={2019},
  url={https://doi.org/10.1145/3368567.3368584},
  doi={10.1145/3368567.3368584}
}

@misc{Meng2024,
 author = {Tao Meng and Ninareh Mehrabi and Palash Goyal and Anil Ramakrishna and Aram Galstyan and Richard Zemel and Kai-Wei Chang and Rahul Gupta and Charith Peris},
 title = {Attribute controlled fine-tuning for large language models: A case study on detoxification},
 year = {2024},
 howpublished = {Amazon Science},
 url = {https://www.amazon.science/publications/attribute-controlled-fine-tuning-for-large-language-models-a-case-study-on-detoxification},
}

@inproceedings{sourabrata-etal-2023-text,
    title = "Text Detoxification as Style Transfer in {E}nglish and {H}indi",
    author = "Mukherjee, Sourabrata  and
      Bansal, Akanksha  and
      Kr. Ojha, Atul  and
      P. McCrae, John  and
      Dusek, Ondrej",
    editor = "D. Pawar, Jyoti  and
      Lalitha Devi, Sobha",
    booktitle = "Proceedings of the 20th International Conference on Natural Language Processing (ICON)",
    month = dec,
    year = "2023",
    address = "Goa University, Goa, India",
    publisher = "NLP Association of India (NLPAI)",
    url = "https://aclanthology.org/2023.icon-1.13/",
    pages = "133--144"
}

@article{dementieva2024overview,
  title={Overview of the multilingual text detoxification task at pan 2024},
  author={Dementieva, Daryna and Moskovskiy, Daniil and Babakov, Nikolay and Ayele, Abinew Ali and Rizwan, Naquee and Schneider, Florian and Wang, Xintong and Yimam, Seid Muhie and Ustalov, Dmitry and Stakovskii, Elisei and Smirnova, Alisa and Elnagar, Ashraf and Mukherjee, Animesh and Panchenko, Alexander},
  journal={CEUR Workshop Proceedings},
  volume={3740},
  pages={2432--2461},
  year={2024},
  url={https://nchr.elsevierpure.com/en/publications/overview-of-the-multilingual-text-detoxification-task-at-pan-2024/}
}

@inproceedings{som-etal-2024-demonstrations,
    title = "Demonstrations Are All You Need: Advancing Offensive Content Paraphrasing using In-Context Learning",
    author = "Som, Anirudh  and
      Sikka, Karan  and
      Gent, Helen  and
      Divakaran, Ajay  and
      Kathol, Andreas  and
      Vergyri, Dimitra",
    editor = "Ku, Lun-Wei  and
      Martins, Andre  and
      Srikumar, Vivek",
    booktitle = "Findings of the Association for Computational Linguistics: ACL 2024",
    month = aug,
    year = "2024",
    address = "Bangkok, Thailand",
    publisher = "Association for Computational Linguistics",
    url = "https://aclanthology.org/2024.findings-acl.749/",
    doi = "10.18653/v1/2024.findings-acl.749",
    pages = "12612--12627"
}

@inproceedings{hartvigsen2022toxigen,
  title={ToxiGen: A Large-Scale Machine-Generated Dataset for Adversarial and Implicit Hate Speech Detection},
  author={Hartvigsen, Thomas and Gabriel, Saadia and Palangi, Hamid and Sap, Maarten and Ray, Dipankar and Kamar, Ece},
  booktitle={Proceedings of the 60th Annual Meeting of the Association for Computational Linguistics (Volume 1: Long Papers)},
  month={may},
  address={Dublin, Ireland},
  publisher={Association for Computational Linguistics},
  pages={3309--3326},
  year={2022},
  url={https://aclanthology.org/2022.acl-long.234/},
  doi={10.18653/v1/2022.acl-long.234}
}

@article{rtplx,
    title={RTP-LX: Can LLMs Evaluate Toxicity in Multilingual Scenarios?},
    volume={39},
    url={https://ojs.aaai.org/index.php/AAAI/article/view/35011},
    DOI={10.1609/aaai.v39i27.35011},
    number={27},
    journal={Proceedings of the AAAI Conference on Artificial Intelligence},
    author={de Wynter, Adrian and Watts, Ishaan and Wongsangaroonsri, Tua and Zhang, Minghui and Farra, Noura and Altıntoprak, Nektar Ege and Baur, Lena and Claudet, Samantha and Gajdušek, Pavel and Gu, Qilong and Kaminska, Anna and Kaminski, Tomasz and Kuo, Ruby and Kyuba, Akiko and Lee, Jongho and Mathur, Kartik and Merok, Petter and Milovanović, Ivana and Paananen, Nani and Paananen, Vesa-Matti and Pavlenko, Anna and Vidal, Bruno Pereira and Strika, Luciano Ivan and Tsao, Yueh and Turcato, Davide and Vakhno, Oleksandr and Velcsov, Judit and Vickers, Anna and Visser, Stéphanie F. and Widarmanto, Herdyan and Zaikin, Andrey and Chen, Si-Qing},
    year={2025},
    month={Apr.},
    pages={27940-27950}
}

@inproceedings{brun2024frenchtoxicityprompts,
  title={FrenchToxicityPrompts: a Large Benchmark for Evaluating and Mitigating Toxicity in French Texts},
  author={Brun, Caroline and Nikoulina, Vassilina},
  booktitle={Proceedings of the Fourth Workshop on Threat, Aggression \& Cyberbullying@ LREC-COLING-2024},
  pages={105--114},
  year={2024},
  url={https://aclanthology.org/2024.trac-1.12/}
}

@inproceedings{kim2024lifetox,
  title={LifeTox: Unveiling Implicit Toxicity in Life Advice},
  author={Kim, Minbeom and Koo, Jahyun and Lee, Hwanhee and Park, Joonsuk and Lee, Hwaran and Jung, Kyomin},
  booktitle={Proceedings of the 2024 Conference of the North American Chapter of the Association for Computational Linguistics: Human Language Technologies (Volume 2: Short Papers)},
  month={jun},
  address={Mexico City, Mexico},
  publisher={Association for Computational Linguistics},
  pages={688--698},
  year={2024},
  url={https://aclanthology.org/2024.naacl-short.60/},
  doi={10.18653/v1/2024.naacl-short.60}
}

@article{duan2025gloss,
  title={GloSS over Toxicity: Understanding and Mitigating Toxicity in LLMs via Global Toxic Subspace},
  author={Duan, Zenghao and Yin, Zhiyi and Shi, Zhichao and Pang, Liang and Jing, Shaoling and Wu, Jiayi and Yan, Yu and Shen, Huawei and Cheng, Xueqi},
  journal={arXiv preprint arXiv:2505.17078},
  year={2025},
  eprint={2505.17078},
  archivePrefix={arXiv},
  primaryClass={cs.CL},
  url={https://arxiv.org/abs/2505.17078},
  doi={10.48550/arXiv.2505.17078}
}

@article{wang2021simple,
  title={Simple text detoxification by identifying a linear toxic subspace in language model embeddings},
  author={Wang, Andrew and Sudhakar, Mohit and Ji, Yangfeng},
  journal={arXiv preprint arXiv:2112.08346},
  year={2021},
  eprint={2112.08346},
  archivePrefix={arXiv},
  primaryClass={cs.CL},
  url={https://arxiv.org/abs/2112.08346},
  doi={10.48550/arXiv.2112.08346}
}

@article{shaik2025redefining,
  title={Redefining Experts: Interpretable Decomposition of Language Models for Toxicity Mitigation},
  author={Shaik, Zuhair Hasan and Mazhar, Abdullah and Srivastava, Aseem and Akhtar, Md Shad},
  journal={arXiv preprint arXiv:2509.16660},
  year={2025},
  eprint={2509.16660},
  archivePrefix={arXiv},
  primaryClass={cs.CL},
  url={https://arxiv.org/abs/2509.16660},
  doi={10.48550/arXiv.2509.16660}
}

@inproceedings{conneau2020unsupervised,
  title={Unsupervised cross-lingual representation learning at scale},
  author={Conneau, Alexis and Khandelwal, Kartikay and Goyal, Naman and Chaudhary, Vishrav and Wenzek, Guillaume and Guzm{\'a}n, Francisco and Grave, Edouard and Ott, Myle and Zettlemoyer, Luke and Stoyanov, Veselin},
  booktitle={Proceedings of the 58th annual meeting of the association for computational linguistics},
  month={jul},
  address={Online},
  publisher={Association for Computational Linguistics},
  pages={8440--8451},
  year={2020},
  url={https://aclanthology.org/2020.acl-main.747/},
  doi={10.18653/v1/2020.acl-main.747}
}

\appendix

\section{Detection and Detoxification Comparisons}

\begin{table*}[t]
\centering
\small
\resizebox{\textwidth}{!}{
\begin{tabular}{p{2.8cm} p{3.3cm} p{3.8cm} p{4.0cm} p{4.1cm}}
\toprule
\textbf{Approach} & \textbf{Typical setup} & \textbf{Strengths} & \textbf{Main failure modes} & \textbf{Representative evidence} \\
\midrule
Cross-lingual classifiers & mBERT/XLM-R or similar encoders trained on labeled toxicity data & Efficient inference; supports many languages with shared representations & Uneven transfer across scripts, dialects, and low-resource languages; annotation bias transfers with labels & Cross-lingual hate-speech detection \citep{ticta2021cross}; dialect and tokenization bias \citep{sap-etal-2019-risk, kanjirangat2025tokenization} \\
\midrule
Translation-to-English pipelines & Translate non-English text, then apply an English detector & Reuses strong English detectors; simple to deploy & Translation artifacts, semantic drift, and toxicity insertion/omission & Cross-lingual toxicity classification via MT \citep{bell2025translate}; toxicity in MT \citep{costa2023toxicity} \\
\midrule
LLM-based detection & Prompt or fine-tune instruction models/guard models for toxicity labels & Flexible label schemas; can use context and rationales & Calibration failures; inconsistent cultural norms; closed-model reproducibility issues & LLM toxicity detection \citep{hu2024toxicity}; guard calibration \citep{liucalibration}; multilingual reasoning guardrails \citep{yang-etal-2025-mrguard} \\
\midrule
Representation probing & Linear probes, subspace analysis, or feature attribution in hidden states & Supports interpretability and targeted mitigation & Correlational unless paired with causal interventions; multilingual transfer remains under-tested & Toxic subspaces \citep{wang2021simple, duan2025gloss}; expert decomposition \citep{shaik2025redefining} \\
\bottomrule
\end{tabular}}
\caption{Comparison of multilingual toxicity detection approaches.}
\label{tab:detection_comparison}
\end{table*}

\begin{table*}[t]
\centering
\small
\resizebox{\textwidth}{!}{
\begin{tabular}{p{2.9cm} p{4.2cm} p{4.2cm} p{4.5cm}}
\toprule
\textbf{Technique} & \textbf{Strengths} & \textbf{Weaknesses} & \textbf{Representative evidence} \\
\midrule
Parallel supervised fine-tuning & Strong task fit when toxic--neutral pairs exist; direct control over rewriting behavior & Expensive parallel data; weak zero-shot transfer; style can flatten & ParaDetox and multilingual extensions \citep{paradetox, multiparadetox, dementieva-etal-2025-multilingual} \\
\midrule
Preference tuning / RLHF / DPO & Can reduce toxic continuations and transfer safety preferences across languages & Transfer varies with representation alignment and language resources; English-heavy reward data can misalign norms & Cross-lingual preference transfer \citep{li-yong-bach-2024-preference, neplenbroek2025cross}; multilingual preference optimization \citep{dang-etal-2024-rlhf} \\
\midrule
Decoding-time steering & Avoids full retraining; can be toggled or tuned at inference time & Needs calibrated classifiers or expert models; may degrade fluency under strong guidance; multilingual evidence remains limited & PPLM \citep{dathathri2020plug}; GeDi \citep{krause2021gedi}; DExperts \citep{liu-etal-2021-dexperts} \\
\midrule
Edit-after-generate & Modular; can combine detection, rewriting, and reranking & Slower; detector errors propagate; translation pivots can lose pragmatics & Toxification reversal \citep{leong2023self}; self-detoxification \citep{ko2024large}; retrieval mitigation \citep{pozzobon-etal-2023-goodtriever} \\
\midrule
Representation editing & Targets internal toxicity features with limited data or parameter updates & Regression risk; limited multilingual evaluation; causal claims require audits & SafeEdit \citep{safedit}; activation engineering \citep{turner2024steeringlanguagemodelsactivation}; SAE steering \citep{goyal2025breaking} \\
\midrule
Multilingual guardrails & Deployment-time prompt/response gating; policy labels can be updated without changing generator & Does not detoxify the generator itself; vulnerable to coverage gaps and adversarial multilingual forms & MultiGuard/OmniGuard \citep{verma2025multiguard}; PolyGuard \citep{kumar2025polyguard}; MrGuard \citep{yang-etal-2025-mrguard}; Qwen3Guard \citep{zhao2025qwen3guard} \\
\bottomrule
\end{tabular}}
\caption{Comparison of multilingual LLM detoxification and moderation techniques.}
\label{tab:detox_techniques}
\end{table*}

\end{document}